\documentclass{article} % For LaTeX2e
\usepackage{iclr2025_conference,times}

\usepackage{amsmath,amsfonts,bm}

\def\eqref#1{equation~\ref{#1}}
\def\1{\bm{1}}

\DeclareMathAlphabet{\mathsfit}{\encodingdefault}{\sfdefault}{m}{sl}
\SetMathAlphabet{\mathsfit}{bold}{\encodingdefault}{\sfdefault}{bx}{n}

\usepackage{enumitem}
\usepackage{hyperref}
\usepackage{url}
\usepackage{graphicx}
\usepackage{subcaption} % For side-by-side figures with (a), (b) labels
\usepackage[most]{tcolorbox} % 'most' loads breakable and other libraries
\usepackage{booktabs}   % For professional-looking tables (\toprule, \midrule, \bottomrule)
\usepackage{multirow}   % To span table cells across multiple rows
\usepackage{amssymb}    % For checkmark (\checkmark) and cross (\times) symbols

\usepackage{pifont}     % For the cross symbol (\ding{55})

\title{MoEs Are Stronger than You Think: \\Hyper-Parallel Inference Scaling with RoE}

\author{
  \begin{tabular}{c}
    Soheil Zibakhsh\footnotemark[1] \footnotemark[2] \textsuperscript{1,2},
    Mohammad Samragh\footnotemark[2] \textsuperscript{1} ,
    Kumari Nishu\textsuperscript{1} , 
    Lauren Hannah\textsuperscript{1} , \\
    Arnav Kundu\textsuperscript{1} , \ and
    Minsik Cho\textsuperscript{1}  \\
    \\
    \textsuperscript{1}Apple, \textsuperscript{2}University of California San Diego
  \end{tabular}%
}

\iclrfinalcopy % Uncomment for camera-ready version, but NOT for submission.
\begin{document}
% --- DEFINE THE FOOTNOTE TEXT HERE ---
\renewcommand{\thefootnote}{\fnsymbol{footnote}}
\footnotetext[1]{Work done during an internship at Apple.}
\footnotetext[2]{Corresponding authors: \texttt{szibakhshshabgahi@ucsd.edu, m\_samraghrazlighi@apple.com}}
\renewcommand{\thefootnote}{\fnsymbol{footnote}}
% ------------------------------------

\maketitle

\begin{abstract}
The generation quality of large language models (LLMs) is often improved by utilizing inference-time sequence-level scaling methods (e.g., Chain-of-Thought). We introduce \textit{hyper-parallel scaling}, a complementary framework that improves prediction quality at the token level. 
Hyper-parallel scaling computes and aggregates multiple output proposals for a single token from the model. 
We implement this concept in Mixture-of-Experts (MoE) models, which we refer to as Roster of Experts (RoE).
RoE is a training-free inference algorithm that turns a single MoE into a dynamic ensemble of MoEs. 
RoE injects controlled stochasticity into the expert routing mechanism, enabling it to sample multiple diverse experts for each token and aggregate their outputs for a more accurate final prediction.
To overcome the computational cost, we introduce an efficient batching strategy and a specialized KV-caching mechanism that minimizes compute and memory overhead. 
For example, RoE enables a 7B MoE model to match the performance of a 10.5B MoE model while using 30\% less compute for inference. These gains are achieved without any fine-tuning of model parameters.
\end{abstract}

\section{Introduction}

\begin{figure}[t]
    \centering
    \includegraphics[width=1.0\textwidth]{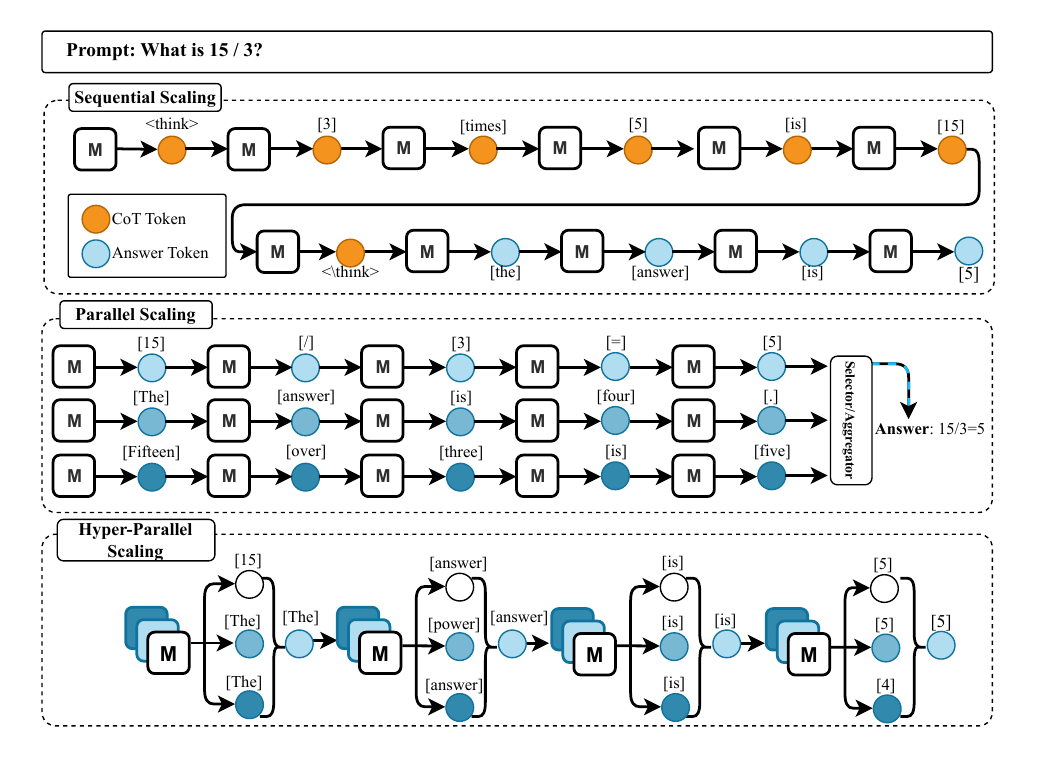}
    \caption{A categorization of inference-time scaling strategies. \textbf{(I) Sequential Scaling:} Enhancing performance by generating longer, structured outputs like a chain of thought \citep{wei2022chain}. \textbf{(II) Parallel Scaling:} Generating multiple token sequences and aggregating them, as in Self-Consistency \citep{wang2022self}. \textbf{(III) Hyper-Parallel Scaling:} A novel paradigm, instantiated by RoE, that aggregates results from diverse internal computation paths on a per-token basis.}
    \label{fig:positioning}
    \vspace{-0.5cm}
\end{figure}

Extensive data and substantial computational resources have fueled recent advancements in language models. 
While the simplest method for generating responses is greedy decoding, the quality of model outputs often requires enhancement at inference time. A growing line of work in this area focuses on test-time scaling, which aims to improve the performance of the sequence generation process. 
% These methods enhance performance by scaling generation at the sequence level, for example through structured reasoning or by ensembling multiple outputs. 
Existing test-time scaling approaches typically fall into two orthogonal categories: sequential scaling, where the model produces longer, more structured outputs (e.g., Chain-of-Thought~\citep{wei2022chain}); and parallel scaling, where multiple independent sequences are generated and then aggregated (e.g., self-consistency~\citep{wang2022self}). The general notion of these categories is marked as ``Sequential Scaling'' and ``Parallel Scaling'' in Figure~\ref{fig:positioning}.

In this paper, we pose an orthogonal question: Can we improve a model’s intrinsic next-token prediction capability by allocating more computation at inference time? In other words, can we increase the model’s internal compute during inference to enhance the quality of every generated token? We refer to this new paradigm as \textit{hyper-parallel scaling}, as shown in Figure~\ref{fig:positioning}. This approach improves generation quality even under the simplest decoding strategy, greedy decoding. To isolate the gains attributable to hyper-parallel scaling, we focus our experiments on evaluating greedy decoding quality throughout the paper.

% Existing approaches for test-time scaling fall iWe pose a more fundamental question: can a model’s intrinsic ability as a next-token predictor be improved independently of the sequence-level generation strategy? 

% Consequently, a significant research effort has focused on navigating the trade-off between computational cost and model performance. Techniques such as quantization~\cite{frantar2022gptq}, caching~\cite{pope2023efficiently}, and speculative decoding~\cite{leviathan2023fast} are employed to reduce inference costs, sometimes accepting a minor degradation in quality.

% In this work, we explore the complementary question: can we improve model performance by allocating more computational resources at inference time? Existing test-time scaling methods enhance performance by scaling the sequence generation process. These approaches fall into two main categories: sequential scaling, where the model generates a longer, more structured output (e.g., Chain-of-Thought~\citep{wei2022chain}); and parallel scaling, where multiple independent sequences are generated and aggregated (e.g., self-consistency~\citep{wang2022self}).We pose a more fundamental question: can a model’s intrinsic ability as a next-token predictor be improved independently of the sequence-level generation strategy? In other words, can we increase the model’s internal compute at inference time, thereby enhancing the quality of every generated token? We refer to this new paradigm as \textit{hyper-parallel scaling} (see Figure~\ref{fig:positioning}).

Hyper-parallel scaling aims to unlock a model’s full potential by increasing the computation allocated to each token at inference time. One way to realize this idea is by introducing controlled variation within each transformer block~\citep{shelmanov2021certain} and recomputing the layer output multiple times. Another approach is to reuse each layer repeatedly in a recurrent manner, thereby increasing computation without adding parameters~\citep{lin2022spin}. While many variants are possible, we focus on sparsely activated Mixture-of-Experts (MoE) models, which provide an ideal architecture for implementing this concept.

Mixture of experts (MoE) models have become a leading solution for frontier large language models~\citep{shazeer2017outrageously,comanici2025gemini,dai2024deepseekmoe}. Since they activate only a fraction of their parameters per forward pass, they naturally raise the central question of hyper-parallel scaling: can the inactive experts be leveraged at inference time to boost performance? Simply increasing the number of active experts does not work, as models are not trained to aggregate information from larger expert sets. To address this, we propose Roster of Experts (RoE), a training-free inference technique that treats a single MoE as a dynamic ensemble. RoE adds controlled stochasticity into the router’s expert selection, runs multiple stochastic forward passes per token, and aggregates the resulting logits into a single, higher-quality prediction, all without model fine-tuning.

As is evident, a naive implementation of RoE would incur substantial redundant computation. We address this by exploiting the overlap across forward passes and merging them into a single batched call to the LLM. Furthermore, we introduce a specialized caching mechanism to reduce the KV-cache size required for RoE generation. 
% For example, simulating 32 variants of OlMoE-7B~\citep{muennighoff2024olmoe} increases runtime by only 12\%, yet the model attains generation quality comparable to a 10.5B model, which is 30\% more expensive to run. 
% For instance, when we simulate running 32 variants of OlMoE-7B, the runtime only increases by 12\%. However, the model achieves generation quality equivalent to a 10.5B model, which is 30\% more expensive to run.
% applying our method to a 7B MoE model while simulating 32 parallel matches the performance of a much larger 10.5B model, while requiring 30\% less latency and 25\% less memory.
% While a naive implementation of this sampling process would incur a prohibitive computational cost, we introduce a specialized caching mechanism that exploits the significant overlap in computation across the different paths. This makes RoE a practical method for unlocking additional performance from pre-trained MoE models. We establish the superior computational efficiency of RoE compared to direct model scaling. For instance, applying our method to a 7B MoE model matches the performance of a much larger 10.5B model, while requiring 30\% less latency and 25\% less memory.
In short, the contributions in this work are as follows:
\begin{itemize}[leftmargin=0.5cm]
    \item We introduce hyper-parallel scaling, a novel inference paradigm that allocates additional compute at test time to diversify a model’s internal computations, thereby improving the quality of each token prediction.
    \item We propose Roster of Experts (RoE), a training-free approach to hyper-parallel scaling in MoE models that ensembles diverse computational paths. RoE leverages Gumbel-Top-K routing to inject controlled stochasticity into expert selection and introduces execution and KV-cache optimizations for efficient inference.
    \item We demonstrate the superior efficiency of RoE compared to conventional model scaling. For instance, we demonstrate that RoE can enhance the OlMoE-7B model~\citep{muennighoff2024olmoe} to achieve the performance of a 10.5B model, with a 30\% latency decrease compared to its larger counterpart. The enhancement requires no model finetuning. % while incurring 30\% less latency and 25\% less memory.
\end{itemize}
    % \item We introduce Gumbel-Top-K routing, a principled mechanism for injecting temperature-controlled stochasticity into the expert selection process.
    % \item We design a novel caching strategy that leverages shared activations across sampled paths, significantly reducing the computational overhead of RoE.

\section{RoE: Hyper Parallel Scaling of Mixture of Experts}
\label{sec:method}
\begin{figure}[tbp]
    \centering
    \includegraphics[width=1.0\textwidth]{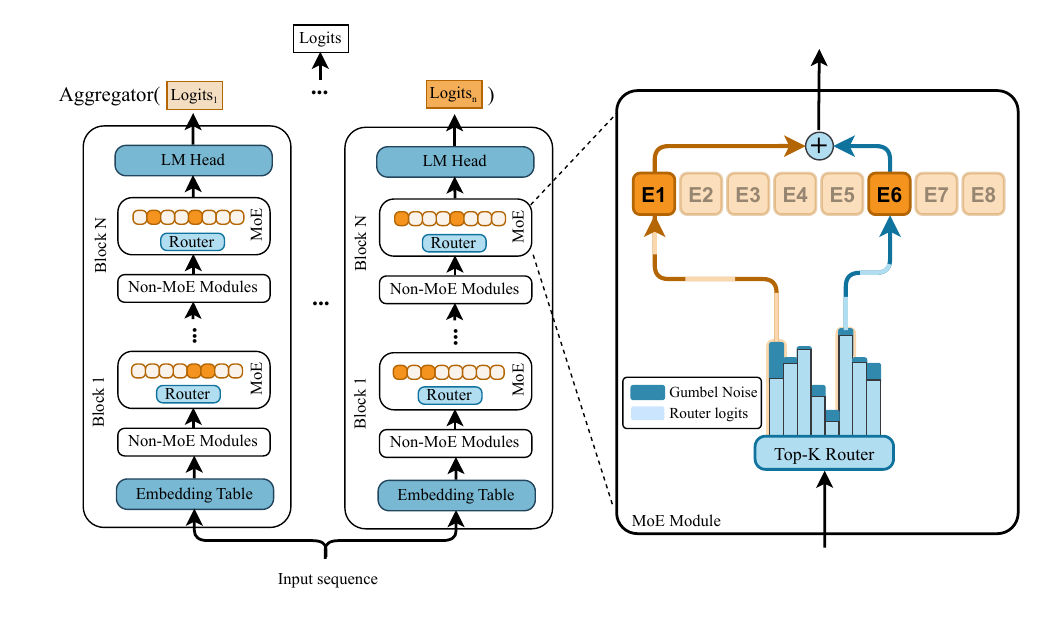}
    \caption{An illustration of the Roster of Experts (RoE) method. \textbf{Left:} For a single input, $n$ distinct experts are sampled by adding stochasticity to the expert routing at each MoE layer, and the resulting output logits are aggregated to form the final prediction. \textbf{Right:} A closer view of a single MoE layer shows $k=2$ active experts (dark orange), where Gumbel noise (dark blue) is added to the router logits, and the top-$k$ experts are selected based on these modified logits.}
    \label{fig:main_idea}
\end{figure}

Roster of Experts (RoE) enhances a pre-trained MoE model's performance by treating it as a dynamic ensemble. In designing RoE, we hypothesize that making controlled variations in routing still yields high-quality predictions. The rationale behind our claim is straightforward: during training, the model already encounters a wide range of expert combinations, so there is no reason not to exploit the same diversity at test time. 
% Our experimental results support this approach.

Given the aforementioned insight, we provide a high-level illustration of RoE in Figure~\ref{fig:main_idea}. At each generation step, the MoE model generates multiple candidate output logits for a single input by sampling diverse expert selections from the model. These outputs are then aggregated to produce a single, more accurate prediction. This process relies on two key components: a stochastic routing mechanism to create diverse paths, and an efficient inference strategy to ensure practicality.

\subsection{Gumbel-Top-K Routing for Path Diversity}

Standard MoE models use deterministic top-k routing, where each token is routed to the $k$ experts with the highest router logits. To generate diverse computational paths, we introduce controlled stochasticity into this selection process using Gumbel-Top-K routing. Given the router logits $R\in \mathbb{R}^E$ for a token over $E$ experts, we perturb them with Gumbel noise before selecting the top-k experts. The indices of the selected experts are given by:
\begin{equation}
\text{Indices} = \text{TopK}(\mathbf{R} + \tau \cdot \mathbf{G}, k)
\end{equation}
where $G$ is a vector of i.i.d. samples from the Gumbel(0, 1) distribution, and $\tau$ is a temperature parameter that controls the degree of stochasticity. When $\tau=0$, this reduces to standard deterministic top-k routing. As $\tau$ increases, the selection becomes more random.

% Explain the Gumbel topK trick and why if \tau is equal to 1 this will be equivalent to sampling from the R distribution with no replacement. If the explanatin is long cite a work that shows this.
This method is a principled way to sample from the distribution implicitly defined by the router logits. The Gumbel-Max trick \citep{gumbel1954statistical} establishes that adding Gumbel noise to logits before an $argmax$ operation is equivalent to sampling from the categorical distribution produced by applying a softmax to the logits. By extension, applying a TopK operation to Gumbel-perturbed logits corresponds to sampling $k$ elements without replacement from the distribution. This ensures that experts with higher router logits remain more likely to be selected even after adding Gumbel noise, preventing the selection from drifting too far from the trained router’s predictions.
Figure~\ref{fig:main_idea} illustrates this mechanism.

% \begin{figure}[tbp]
% \centering
% \includegraphics[width=1.0\textwidth]{figures/Gumbel-topk-1.pdf}
%     \caption{A comparison of deterministic Top-K routing and our proposed Gumbel-Top-K routing. While standard Top-K always selects the experts with the highest logits (left), Gumbel-Top-K (right) introduces temperature-controlled stochasticity, enabling the selection of diverse, high-probability experts.}
% \label{fig:gumbel_router}
% \end{figure}
\subsection{Choosing the Gumbel Temperature}
The Gumbel temperature, $\tau$, controls the degree of stochasticity in expert routing. We treat it as a layer-specific hyperparameter, defining a temperature vector $\boldsymbol{\tau} = \{\tau_i\}_{i \in \mathcal{L}_{MoE}}$, where $\mathcal{L}_{MoE}$ is the set of MoE layers. 
Setting a small value of $\tau$ keeps router selections nearly unchanged, reducing expert diversity per sample and making the next-token prediction closely match the underlying MoE. In contrast, an excessively large $\tau$ introduces too much randomness in expert selection, degrading prediction quality. Appendix~\ref{app:temp_analysis} illustrates how task performance varies as the temperature increases.
Selecting the optimal $\boldsymbol{\tau}$ presents a hyperparameter optimization problem, balancing the potential performance gain against the cost of the search. The primary challenge is the search space, which grows exponentially with the number of MoE layers, rendering an exhaustive search infeasible. A practical search strategy therefore requires an efficient optimization algorithm and a carefully chosen validation metric.

\subsubsection{Optimization Metric}
We consider two metrics for guiding the hyperparameter search: validation perplexity and task-specific accuracy.

\textbf{Validation Perplexity (PPL)} is computationally inexpensive, as it only requires a single forward pass over the validation set. However, PPL is an indirect measure of generative reasoning. A low PPL indicates that the model assigns a high probability to a ground-truth sequence, but does not guarantee that the model can generate a correct solution independently.

\textbf{Validation Accuracy} directly measures the model's ability to solve the task, making it a more faithful metric for generative performance. The main drawback is its high computational cost, as it requires generating a full solution for each validation example. This cost can make it prohibitive for large-scale hyperparameter searches.

\subsubsection{Search Strategy}
Given the cost of each evaluation, Bayesian optimization methods are well-suited for this task. In our experiments (Section~\ref{sec:results}), we employ the Tree-structured Parzen Estimator (TPE)~\cite{watanabe2023tree} via the Optuna framework~\cite{akiba2019optuna}. To make the search tractable for models with many MoE layers, we introduce two heuristics to prune the vast search space, based on empirical observations:

\paragraph{1. Apply RoE to middle layers only.} We hypothesize that the initial and final layers of a transformer are more sensitive to routing perturbations. The initial layers process raw token embeddings, while the final layers consolidate information for the output prediction. Our experiments show preference of the optimizer to reduce the stochasticity of initial and final layers. We therefore constrain the search by setting the temperature to zero ($\tau_i=0$) for the first and last $k$ MoE layers, applying RoE only to the intermediate ones.

\paragraph{2. Bound the temperature range.} We empirically observe that temperature values above 0.5 introduce excessive routing noise, which consistently harms model performance. Consequently, we restrict the search space for each non-zero temperature to the range $[0, 0.5]$.

\subsection{Efficient RoE Inference}

A naive implementation of RoE, which performs $n$ independent forward passes, would incur a prohibitive n-fold increase in computation. We introduce two optimization techniques that drastically reduce this overhead.

First, we take advantage of the batched parallel processing capabilities of modern accelerators, such as GPUs. The latency of a forward pass grows sub-linearly with the batch size due to hardware-level parallelization. By processing the $n$ samples for a single token generation step as a batch, we can significantly reduce the wall-clock time compared to $n$ sequential runs.

We demonstrate in Section~\ref{subsubsec:compute_cost} that Key-Value (KV) caching significantly reduces the computational overhead of sequence generation with RoE.
However, batched inference alone does not solve the significant memory overhead introduced by the KV-cache in autoregressive decoding. Since the $n$ samples follow different computational paths, their hidden states diverge. A naive implementation would require maintaining $n$ separate KV caches, one for each sample's unique history. This scales both memory and computation linearly with $n$, quickly becoming intractable for long sequences.

To address this, we introduce a novel caching strategy named \textbf{Clean Cache}. Our key insight is that sufficient output diversity can be achieved by applying stochastic routing only for the current token being generated, while all samples share a common KV cache derived from a single, deterministic history. We implement this efficiently within our batched approach by setting the routing temperature $\tau$ for the first sample (batch index 0) to zero, making it the ``clean'' path. We then store and reuse the KV cache from this single clean sample across all other samples in the batch. As a result, the memory footprint of the Clean Cache is identical to that of a single sample's KV cache, incurring no extra overhead compared to regular caching. This localizes the additional cost of RoE entirely to the batched forward pass of the current token, making it a practical inference-time strategy.
\section{Experiments}
\label{sec:results}

\subsection{Experimental Setup}

\paragraph{Models and Benchmarks.}
We evaluate RoE on three instruction-tuned MoE models: OLMoE-1B-7B-Instruct~\citep{muennighoff2024olmoe}, Mixtral-8x7B-Instruct-v0.1~\citep{jiang2024mixtral}, and GPT-OSS-20B~\citep{openai2025gptoss120bgptoss20bmodel}. Our evaluation spans three domains: mathematical reasoning (GSM8K~\citep{gsm2021training}, SVAMP~\citep{patel2021svamp}, AddSub~\citep{hosseini2014addsub}, SingleEQ~\citep{koncel2015singleeq}, MultiArith~\citep{imani2023multiarith}), commonsense reasoning (ARC-Easy, ARC-Challenge~\citep{clark2018arc}, OpenBookQA~\citep{mihaylov2018openbookqa}, Social-I-QA~\citep{sap2019socialiqa}, Hellaswag~\citep{zellers2019hellaswag}), and code generation (HumanEval~\citep{chen2021humaneval}, HumanEvalPlus~\citep{liu2023humanevalplus}).

\paragraph{Baseline and Evaluation Strategy.}
Our primary goal is to isolate the performance gains directly attributable to RoE's modification of the model's internal computational pathways. Unlike test-time methods that treat the model as a black box and ensemble outputs from multiple generation runs (e.g., self-consistency), RoE enhances the model's backbone directly. These two classes of methods are orthogonal and can be used in conjunction. Therefore, to ensure a clean and direct measurement of RoE's impact, we use the standard, unmodified MoE inference as our sole baseline. We employ greedy decoding for all experiments, ensuring that any observed improvements stem from RoE itself, not from interactions with complex decoding strategies.

\paragraph{RoE Configuration and Tuning.}
We tune the per-layer noise temperatures for RoE on each task’s validation set using the Tree-structured Parzen Estimator (TPE)~\citep{watanabe2023tree} implemented in Optuna~\citep{akiba2019optuna}. All reported results are evaluated on the \textbf{test sets} after tuning, ensuring no data contamination from the \textbf{validation sets}. For computational efficiency, we optimize validation perplexity for math tasks and validation accuracy for commonsense and code tasks. We employ our `clean-cache` implementation (Section~\ref{sec:method}) for OLMoE and the standard cache for other models. The tuning budget and key RoE hyperparameters are summarized in the appendix (Table~\ref{tab:experiment-setup}, while the complete per-layer temperature profiles are visualized in the appendix (Figure~\ref{fig:heatmap}).

\paragraph{Implementation Details.}
Experiments use NVIDIA A100 80GB GPUs. RoE predictions are aggregated by probability averaging. For commonsense tasks, we follow~\citet{hu2023llmadapters} and select the answer with the highest log-probability. Results are averaged over 5 seeds. Inference latency is measured as the wall-clock time for generating 128 tokens. We report exact-match accuracy for all tasks and pass@1 additionally for code tasks.

\begin{figure}[t]
    \centering
    \includegraphics[width=1.0\textwidth]{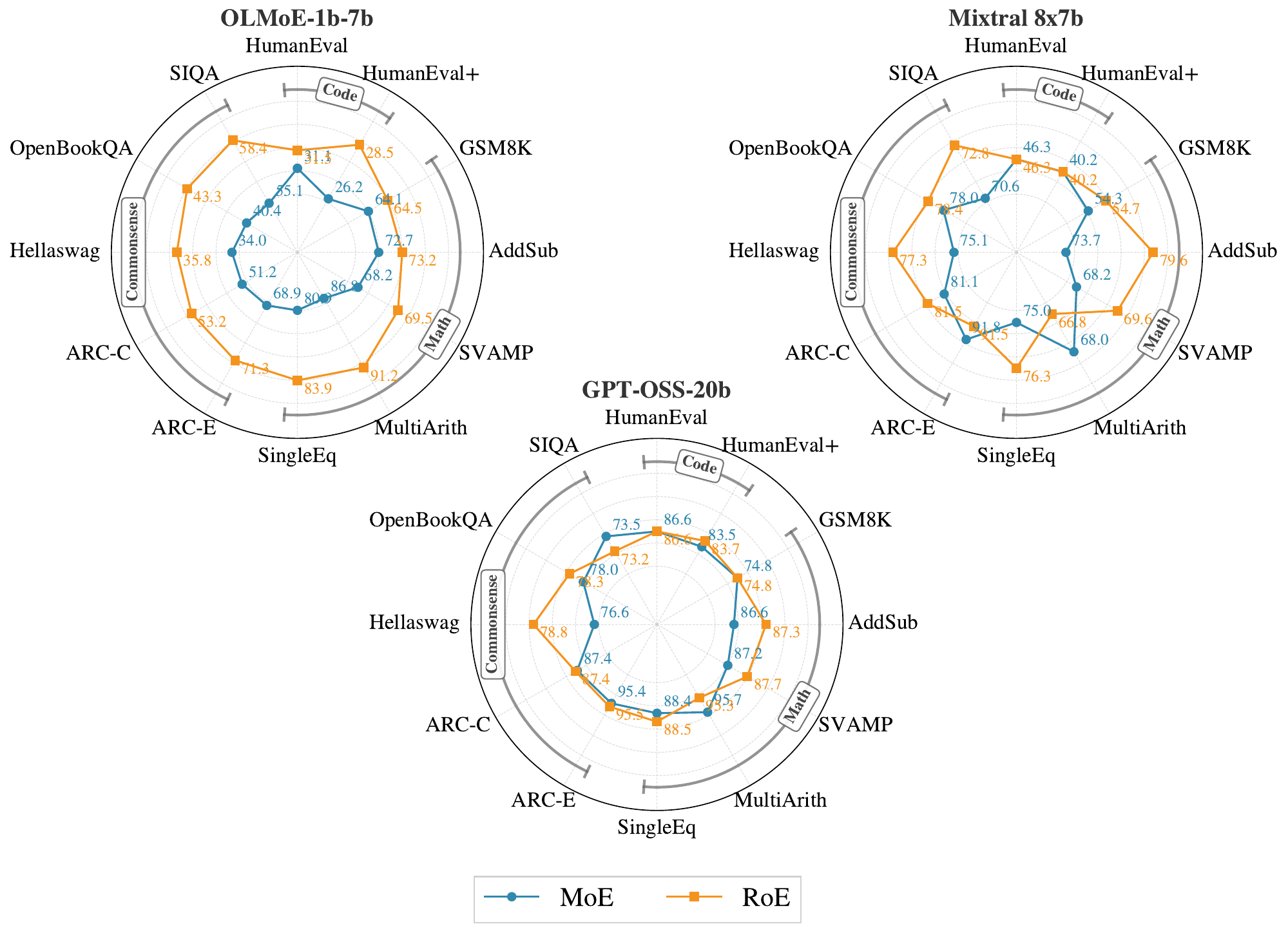}
    \caption{Performance comparison of base MoE models and RoE on five mathematical, five commonsense, and two code benchmarks. Accuracy is measured by exact match (except for HE and HE+, which use pass@1). Results are averaged over five random seeds. Axes are normalized to $(\text{min}-1, \text{max}+1)$ for visualization.}
    \label{fig:main_result}
\end{figure}

\subsection{Main Results}
Figure~\ref{fig:main_result} presents the performance of RoE applied to the three MoE models across our suite of benchmarks. As is evident, RoE consistently improves performance across nearly all tasks and model scales, demonstrating its broad effectiveness as a post-hoc enhancement strategy.

The performance gains are most pronounced for OLMoE, the smallest and weakest base model, where RoE substantially lifts accuracy across all benchmark categories. We hypothesize that by diversifying computational paths, RoE unlocks latent capabilities inaccessible to a single, static routing configuration. This suggests the diversification is most impactful for models with more initial headroom for improvement.

However, the improvements are not uniform. For math benchmarks, we optimized RoE's hyperparameters for validation set perplexity. While RoE consistently lowered perplexity (see Figure~\ref{fig:opt_history_mixral} in the appendix), this did not always translate to higher generative accuracy. For instance, on MultiArith, RoE did not improve the accuracy of Mixtral or GPT-OSS. This result underscores the known disconnect between perplexity and generative reasoning performance, suggesting that tuning RoE directly on task-specific metrics could yield further gains where computationally feasible.

Finally, we observe a ceiling effect: as a model's baseline performance on a benchmark approaches saturation, the potential for improvement diminishes. For instance, the GPT-OSS model already achieves 95.7\% on MultiArith, leaving minimal room for any method to provide substantial gains.

A key advantage of RoE is its applicability to open-ended generation, where parallel scaling methods, like majority voting over complete outputs, are infeasible. By aggregating logits at each token-generation step, RoE enhances the performance of open-ended tasks such as code generation, as evident in Figure~\ref{fig:main_result}.

% \subsection{Ablations}

\subsection{Computational Overhead}
\label{subsubsec:compute_cost}

\begin{figure}[t!]
    \centering
    \includegraphics{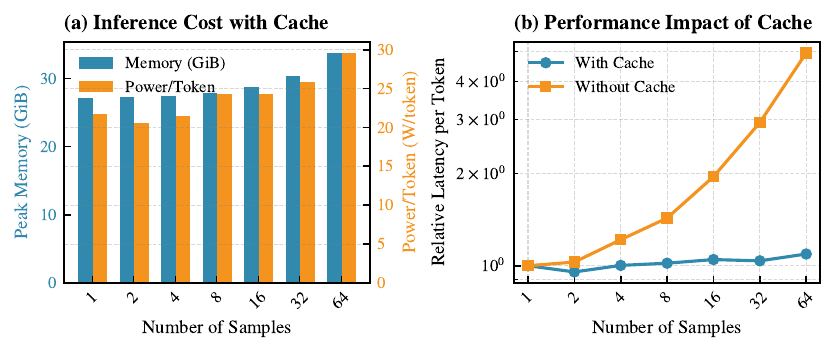}
    \caption{
        Impact of caching on RoE performance. 
        \textbf{(a)}~Resource usage with caching enabled. Peak memory (blue, left axis) and power per token (orange, right axis) show a modest increase as the sample count grows. 
        \textbf{(b)}~Latency comparison with and without caching. Without caching, latency per token rises exponentially, highlighting the necessity of caching for scalable RoE inference.
    }

    \label{fig:caching_performance}
\end{figure}

We evaluate the computational overhead of RoE in terms of GPU memory footprint, power usage, and throughput (tokens/second). Our benchmark uses the OLMoE model to generate solutions for the first 100 problems from the GSM8k benchmark~\citep{gsm2021training} on a single NVIDIA A100 GPU. To isolate the overhead of our method and ensure a fair comparison, we set the generation temperature and $\tau$ to 0, which produces identical output sequences across all runs.

Figure~\ref{fig:caching_performance}(a) illustrates the relative increase in inference cost for RoE with our caching mechanism. The resource requirements grow modestly with the number of samples. For instance, using 64 samples increases the memory footprint by only 12\% and power consumption by 20\%. These results show that the additional computational cost is manageable and not prohibitive.

Figure~\ref{fig:caching_performance}(b)  underscores the necessity of our caching scheme. By comparing RoE with and without caching, we observe that our approach drastically mitigates the increase in power consumption and latency. Without caching, the cost of sampling multiple paths would be impractical. These findings confirm that RoE offers a practical trade-off, enhancing model quality for a moderate and controllable increase in computational demand.

\subsection{Efficiency Analysis: RoE vs. Model Scaling}

\begin{figure}[h!]
    \centering
    \includegraphics{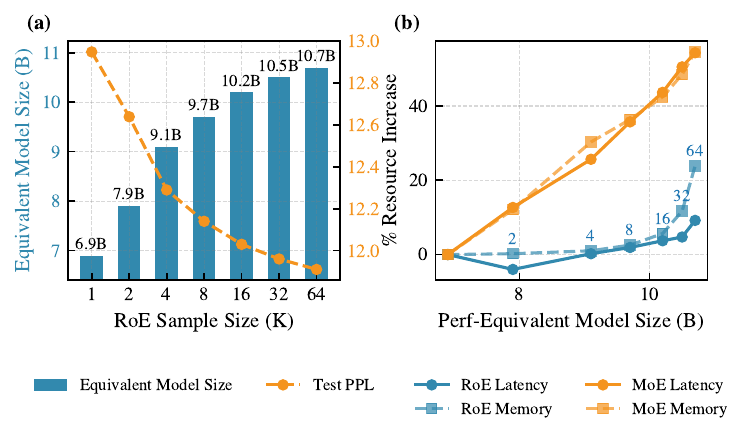}
    \caption{
        Performance and efficiency analysis of RoE. 
        \textbf{(a)}~The performance of RoE applied to OLMoE-7B, measured in terms of an equivalent standard MoE model size. Performance is evaluated using perplexity on the WikiText-103 test set.
        \textbf{(b)}~Comparison of the relative increase in latency and memory for RoE (blue) versus scaling up to an equivalently performing MoE model (orange). The numbers on the blue curve indicate the RoE sample size $K$.
    }
    \label{fig:compute_size}
\end{figure}

RoE enhances a model's performance, making it comparable to a larger version of the base model. To quantify this improvement, we measure the \textit{equivalent model size} that a standard MoE would need to match the performance of RoE for a given number of samples $K$. This allows us to directly compare the efficiency of RoE against the cost of simply using a larger MoE. We estimate this equivalent size by applying the scaling laws from \citet{kaplan2020scaling}, using test perplexity on the WikiText-103 dataset \citep{merity2016pointer} as the performance metric. For each sample size $K$, we tune the RoE routing temperatures across all layers by optimizing perplexity on the WikiText-103 validation set over 50 trials with a TPE optimizer. We then measure the test perplexity of the best-performing temperature setting on the validation set. As shown in Figure~\ref{fig:compute_size}(a), the effective model size grows monotonically with $K$, although the gains diminish as the sample size increases.

Figure~\ref{fig:compute_size}(b) compares the computational cost of RoE (blue curve) with the cost of using a larger, equivalently performing standard MoE model (orange curve). The results show that RoE is substantially more efficient in terms of both latency and memory. For instance, applying RoE with $K=32$ to the OLMoE-7B model yields performance comparable to that of a 10.5B OLMoE model. Notably, this configuration reduces memory overhead by 25\% and per-token latency by 30\% relative to the larger model, underscoring the efficiency of RoE as an alternative to conventional model scaling.

\section{Related Work}

% \begin{figure}[t]
%     \centering
%     \includegraphics[width=1.0\textwidth]{figures/hyper_parallel_option2.pdf}
%     \caption{A categorization of inference-time scaling strategies. \textbf{(I) Sequential Scaling:} Enhancing performance by generating longer, structured outputs like a chain of thought \citep{wei2022chain}. \textbf{(II) Parallel Scaling:} Generating multiple independent reasoning paths and aggregating the results, as in Self-Consistency \citep{wang2022self}. \textbf{(III) Hyper-Parallel Scaling:} A novel paradigm, instantiated by RoE, that aggregates results from diverse internal computation paths on a per-token basis to improve the underlying model's predictive power.}
%     \label{fig:positioning}
% \end{figure}

% \mohammad{figure 5: looks great. A few comments: 1- for sequential scaling, add the <think> and </think> tokens before and after then CoT tokens. 2- For Parallel Scaling, shouldnt' the response be a whole sequence? why is it only one token? 3- For all cases, please remove the "+" that appears the token. It appears as if "+" is part of the token, especially since the context is math. 4-make sure all three subfigures have the same height. 5- make the [M] rectangle smaller to save space.}

Our work, Roster of Experts (RoE), introduces a method for trading additional inference-time compute for improved model performance.  
Several strategies improve model performance by increasing computation at inference time. Inspired by \citet{mirtaheri2025let} we categorize them into three paradigms, as illustrated in Figure~\ref{fig:positioning}.

\noindent{\bf Sequential Scaling:}
Sequential scaling methods improve reasoning quality by prompting the model to generate longer, more structured intermediate steps. The seminal work in this area is Chain-of-Thought (CoT) prompting \citep{wei2022chain, nye2021show}, which elicits step-by-step reasoning. In this category, we ask the model to think, question it self and do a step by step analysis \citep{openai_gpt5, kojima2022large}. This approach has been extended by a large body of subsequent work that explores more complex reasoning structures, such as Tree-of-Thoughts \citep{yao2023tree}, at the cost of increased sequential generation steps. The language model needs to be trained with CoT prompts to be able to utilize this capability during inference.

\noindent{\bf Parallel Scaling:}
Parallel scaling generates multiple independent outputs for the same prompt and aggregates them to produce a final, more robust answer \citep{brown2024large}. The simplest technique to mention for this category is beam search~\cite{freitag2017beam}, where multiple sequences of tokens are evolved at the inference time and the best sequence is chosen as the response. Another prominent example is Self-Consistency (SC) \citep{wang2022self}, which samples diverse reasoning paths using stochastic decoding (e.g., non-zero temperature) and selects the most frequent answer via majority vote. While highly effective, the standard voting mechanism limits SC to tasks with easily verifiable answers, such as math or multiple-choice questions. Recent work aims to extend this paradigm to open-ended generation tasks by developing more sophisticated aggregation strategies that do not rely on a single, extractable answer and instead aggregate on a sequence level \citep{chen2023universal, taubenfeld2025confidence, wang2024soft}. 

\noindent{\bf Hyper-Parallel Scaling:}
We introduce \textit{hyper-parallel scaling} as a distinct, third paradigm. Unlike sequential and parallel methods that scale the \textit{output generation process}, hyper-parallel scaling diversifies the \textit{internal computation paths} within the model for a single token prediction. RoE actualizes this concept by treating a single MoE model as an ensemble of subnetworks. By sampling different sets of active experts for each forward pass, RoE aggregates multiple internal predictions to produce a single, higher-quality output distribution for the next token.
This approach is orthogonal to and can be combined with both sequential and parallel scaling. 

A related direction was explored by \citet{geiping2025scaling}, who designed a special recurrent architecture allowing for variable-depth inference. However, their method requires a model to be specifically pretrained for this capability. In contrast, RoE is a training-free strategy that can be applied post-hoc to any pretrained MoE model, offering a practical way to unlock enhanced performance from existing artifacts dynamically, and on demand.

% \subsection{Efficient LLM Inference}
% \mohammad{let's remove this section. I do not think it is relevant. we made enough references to this category in the introduction.}
% A significant body of research focuses on reducing the computational cost of large language model (LLM) inference. This is often achieved by compressing the model post-training. \textbf{Quantization} methods reduce the numerical precision of model weights (e.g., from 32-bit floats to 8-bit integers), thereby decreasing memory footprint and accelerating computation, often with a minimal drop in performance \citep{dettmers2022gpt3, frantar2022gptq, lin2024awq}. \textbf{Pruning} techniques identify and remove redundant weights from the network, creating smaller, sparser models that are faster to run \citep{frantar2023sparsegpt, sun2023simple}. Another key optimization is \textbf{KV Caching}, which stores the keys and values of past tokens to avoid redundant computations during auto-regressive decoding, trading increased memory for a significant reduction in latency.

\section{Conclusion}
\label{sec:conclusion}

We introduced hyper-parallel scaling, a novel inference paradigm for improving model quality that is orthogonal to existing sequence-level approaches. Instead of generating multiple sequences, this paradigm enhances a model's intrinsic next-token prediction by diversifying its internal computation. We presented Roster of Experts (RoE), a training-free realization of this concept for Mixture-of-Experts (MoE) models. RoE treats a single MoE model as a dynamic ensemble, leveraging controlled stochasticity in the expert routing to activate diverse experts and aggregate their outputs into a more robust prediction.
Crucially, we demonstrated that through efficient batching and caching, the computational overhead of this method is minimal. Our results show that RoE enables a model to achieve the performance of a substantially larger counterpart with only a minor increase in inference cost. This provides practitioners with a powerful tool to dynamically trade inference-time compute for higher quality from a single, pre-trained model.

\section{Future Work}
\label{sec:future_work}

Our work on Roster of Experts (RoE) opens several promising avenues for future research. We highlight three key directions:

\begin{itemize}
    \item \textbf{Generalizing Hyper-Parallel Scaling:} A key direction is to extend hyper-parallel scaling beyond MoE models. Unlike sequence-level scaling methods that are specific to auto-regressive tasks, hyper-parallel scaling is domain-agnostic. This is particularly relevant for modalities like vision, audio, and video, where test-time performance scaling is largely unexplored. Hyper-parallel scaling thus offers a promising path to achieve dynamic performance trade-offs in these domains.
    \item \textbf{Advanced Noise Injection:} Developing more sophisticated noise schemes, such as adaptive or input-conditioned noise, to achieve more effective diversification of expert selection and further improve performance.
    \item \textbf{Adaptive Computation:} Investigating strategies to dynamically adjust RoE's computational budget, for instance by varying the ensemble size based on token difficulty, to optimize the performance-cost trade-off.
    \item \textbf{RoE-Aware Training:} Incorporating stochastic routing into the pre-training or fine-tuning process to create models explicitly optimized for RoE, potentially yielding greater gains in efficiency and quality.
\end{itemize}

\bibliography{iclr2025_conference}

\begin{thebibliography}{38}
\providecommand{\natexlab}[1]{#1}
\providecommand{\url}[1]{\texttt{#1}}
\expandafter\ifx\csname urlstyle\endcsname\relax
  \providecommand{\doi}[1]{doi: #1}\else
  \providecommand{\doi}{doi: \begingroup \urlstyle{rm}\Url}\fi

\bibitem[Akiba et~al.(2019)Akiba, Sano, Yanase, Ohta, and Koyama]{akiba2019optuna}
Takuya Akiba, Shotaro Sano, Toshihiko Yanase, Takeru Ohta, and Masanori Koyama.
\newblock Optuna: A next-generation hyperparameter optimization framework.
\newblock In \emph{Proceedings of the 25th ACM SIGKDD international conference on knowledge discovery \& data mining}, pp.\  2623--2631, 2019.

\bibitem[Brown et~al.(2024)Brown, Juravsky, Ehrlich, Clark, Le, R{\'e}, and Mirhoseini]{brown2024large}
Bradley Brown, Jordan Juravsky, Ryan Ehrlich, Ronald Clark, Quoc~V Le, Christopher R{\'e}, and Azalia Mirhoseini.
\newblock Large language monkeys: Scaling inference compute with repeated sampling.
\newblock \emph{arXiv preprint arXiv:2407.21787}, 2024.

\bibitem[Chen et~al.(2021)Chen, Tworek, Jun, Yuan, Pinto, Kaplan, Edwards, Burda, Joseph, Brockman, et~al.]{chen2021humaneval}
Mark Chen, Jerry Tworek, Heewoo Jun, Qiming Yuan, Henrique Ponde De~Oliveira Pinto, Jared Kaplan, Harri Edwards, Yuri Burda, Nicholas Joseph, Greg Brockman, et~al.
\newblock Evaluating large language models trained on code.
\newblock \emph{arXiv preprint arXiv:2107.03374}, 2021.

\bibitem[Chen et~al.(2023)Chen, Aksitov, Alon, Ren, Xiao, Yin, Prakash, Sutton, Wang, and Zhou]{chen2023universal}
Xinyun Chen, Renat Aksitov, Uri Alon, Jie Ren, Kefan Xiao, Pengcheng Yin, Sushant Prakash, Charles Sutton, Xuezhi Wang, and Denny Zhou.
\newblock Universal self-consistency for large language model generation.
\newblock \emph{arXiv preprint arXiv:2311.17311}, 2023.

\bibitem[Clark et~al.(2018)Clark, Cowhey, Etzioni, Khot, Sabharwal, Schoenick, and Tafjord]{clark2018arc}
Peter Clark, Isaac Cowhey, Oren Etzioni, Tushar Khot, Ashish Sabharwal, Carissa Schoenick, and Oyvind Tafjord.
\newblock Think you have solved question answering? try arc, the ai2 reasoning challenge.
\newblock \emph{arXiv preprint arXiv:1803.05457}, 2018.

\bibitem[Cobbe et~al.(2021)Cobbe, Kosaraju, Bavarian, Chen, Jun, Kaiser, Plappert, Tworek, Hilton, Nakano, et~al.]{gsm2021training}
Karl Cobbe, Vineet Kosaraju, Mohammad Bavarian, Mark Chen, Heewoo Jun, Lukasz Kaiser, Matthias Plappert, Jerry Tworek, Jacob Hilton, Reiichiro Nakano, et~al.
\newblock Training verifiers to solve math word problems.
\newblock \emph{arXiv preprint arXiv:2110.14168}, 2021.

\bibitem[Comanici et~al.(2025)Comanici, Bieber, Schaekermann, Pasupat, Sachdeva, Dhillon, Blistein, Ram, Zhang, Rosen, et~al.]{comanici2025gemini}
Gheorghe Comanici, Eric Bieber, Mike Schaekermann, Ice Pasupat, Noveen Sachdeva, Inderjit Dhillon, Marcel Blistein, Ori Ram, Dan Zhang, Evan Rosen, et~al.
\newblock Gemini 2.5: Pushing the frontier with advanced reasoning, multimodality, long context, and next generation agentic capabilities.
\newblock \emph{arXiv preprint arXiv:2507.06261}, 2025.

\bibitem[Dai et~al.(2024)Dai, Deng, Zhao, Xu, Gao, Chen, Li, Zeng, Yu, Wu, et~al.]{dai2024deepseekmoe}
Damai Dai, Chengqi Deng, Chenggang Zhao, RX~Xu, Huazuo Gao, Deli Chen, Jiashi Li, Wangding Zeng, Xingkai Yu, Yu~Wu, et~al.
\newblock Deepseekmoe: Towards ultimate expert specialization in mixture-of-experts language models.
\newblock \emph{arXiv preprint arXiv:2401.06066}, 2024.

\bibitem[Freitag \& Al-Onaizan(2017)Freitag and Al-Onaizan]{freitag2017beam}
Markus Freitag and Yaser Al-Onaizan.
\newblock Beam search strategies for neural machine translation.
\newblock \emph{arXiv preprint arXiv:1702.01806}, 2017.

\bibitem[Geiping et~al.(2025)Geiping, McLeish, Jain, Kirchenbauer, Singh, Bartoldson, Kailkhura, Bhatele, and Goldstein]{geiping2025scaling}
Jonas Geiping, Sean McLeish, Neel Jain, John Kirchenbauer, Siddharth Singh, Brian~R Bartoldson, Bhavya Kailkhura, Abhinav Bhatele, and Tom Goldstein.
\newblock Scaling up test-time compute with latent reasoning: A recurrent depth approach.
\newblock \emph{arXiv preprint arXiv:2502.05171}, 2025.

\bibitem[Gumbel(1954)]{gumbel1954statistical}
Emil~Julius Gumbel.
\newblock \emph{Statistical theory of extreme values and some practical applications: a series of lectures}, volume~33.
\newblock US Government Printing Office, 1954.

\bibitem[Hosseini et~al.(2014)Hosseini, Hajishirzi, Etzioni, and Kushman]{hosseini2014addsub}
Mohammad~Javad Hosseini, Hannaneh Hajishirzi, Oren Etzioni, and Nate Kushman.
\newblock Learning to solve arithmetic word problems with verb categorization.
\newblock In \emph{Proceedings of the 2014 conference on empirical methods in natural language processing (EMNLP)}, pp.\  523--533, 2014.

\bibitem[Hu et~al.(2023)Hu, Lan, Wang, Xu, Lim, Lee, Bing, and Poria]{hu2023llmadapters}
Zhiqiang Hu, Yihuai Lan, Lei Wang, Wanyu Xu, Ee-Peng Lim, Roy Ka-Wei Lee, Lidong Bing, and Soujanya Poria.
\newblock Llm-adapters: An adapter family for parameter-efficient fine-tuning of large language models.
\newblock \emph{arXiv preprint arXiv:2304.01933}, 2023.

\bibitem[Imani et~al.(2023)Imani, Du, and Shrivastava]{imani2023multiarith}
Shima Imani, Liang Du, and Harsh Shrivastava.
\newblock Mathprompter: Mathematical reasoning using large language models.
\newblock \emph{arXiv preprint arXiv:2303.05398}, 2023.

\bibitem[Jiang et~al.(2024)Jiang, Sablayrolles, Roux, Mensch, Savary, Bamford, Chaplot, Casas, Hanna, Bressand, et~al.]{jiang2024mixtral}
Albert~Q Jiang, Alexandre Sablayrolles, Antoine Roux, Arthur Mensch, Blanche Savary, Chris Bamford, Devendra~Singh Chaplot, Diego de~las Casas, Emma~Bou Hanna, Florian Bressand, et~al.
\newblock Mixtral of experts.
\newblock \emph{arXiv preprint arXiv:2401.04088}, 2024.

\bibitem[Kaplan et~al.(2020)Kaplan, McCandlish, Henighan, Brown, Chess, Child, Gray, Radford, Wu, and Amodei]{kaplan2020scaling}
Jared Kaplan, Sam McCandlish, Tom Henighan, Tom~B Brown, Benjamin Chess, Rewon Child, Scott Gray, Alec Radford, Jeffrey Wu, and Dario Amodei.
\newblock Scaling laws for neural language models.
\newblock \emph{arXiv preprint arXiv:2001.08361}, 2020.

\bibitem[Kojima et~al.(2022)Kojima, Gu, Reid, Matsuo, and Iwasawa]{kojima2022large}
Takeshi Kojima, Shixiang~Shane Gu, Machel Reid, Yutaka Matsuo, and Yusuke Iwasawa.
\newblock Large language models are zero-shot reasoners.
\newblock \emph{Advances in neural information processing systems}, 35:\penalty0 22199--22213, 2022.

\bibitem[Koncel-Kedziorski et~al.(2015)Koncel-Kedziorski, Hajishirzi, Sabharwal, Etzioni, and Ang]{koncel2015singleeq}
Rik Koncel-Kedziorski, Hannaneh Hajishirzi, Ashish Sabharwal, Oren Etzioni, and Siena~Dumas Ang.
\newblock Parsing algebraic word problems into equations.
\newblock \emph{Transactions of the Association for Computational Linguistics}, 3:\penalty0 585--597, 2015.

\bibitem[Lin et~al.(2022)Lin, Prabhu, Merth, Mehta, Ranjan, Horton, and Rastegari]{lin2022spin}
Chien-Yu Lin, Anish Prabhu, Thomas Merth, Sachin Mehta, Anurag Ranjan, Maxwell Horton, and Mohammad Rastegari.
\newblock Spin: an empirical evaluation on sharing parameters of isotropic networks.
\newblock In \emph{European conference on computer vision}, pp.\  553--568. Springer, 2022.

\bibitem[Liu et~al.(2023)Liu, Xia, Wang, and Zhang]{liu2023humanevalplus}
Jiawei Liu, Chunqiu~Steven Xia, Yuyao Wang, and Lingming Zhang.
\newblock Is your code generated by chatgpt really correct? rigorous evaluation of large language models for code generation.
\newblock \emph{Advances in Neural Information Processing Systems}, 36:\penalty0 21558--21572, 2023.

\bibitem[Merity et~al.(2016)Merity, Xiong, Bradbury, and Socher]{merity2016pointer}
Stephen Merity, Caiming Xiong, James Bradbury, and Richard Socher.
\newblock Pointer sentinel mixture models, 2016.

\bibitem[Mihaylov et~al.(2018)Mihaylov, Clark, Khot, and Sabharwal]{mihaylov2018openbookqa}
Todor Mihaylov, Peter Clark, Tushar Khot, and Ashish Sabharwal.
\newblock Can a suit of armor conduct electricity? a new dataset for open book question answering.
\newblock \emph{arXiv preprint arXiv:1809.02789}, 2018.

\bibitem[Mirtaheri et~al.(2025)Mirtaheri, Edelman, Jelassi, Malach, and Boix-Adsera]{mirtaheri2025let}
Parsa Mirtaheri, Ezra Edelman, Samy Jelassi, Eran Malach, and Enric Boix-Adsera.
\newblock Let me think! a long chain-of-thought can be worth exponentially many short ones.
\newblock \emph{arXiv preprint arXiv:2505.21825}, 2025.

\bibitem[Muennighoff et~al.(2024)Muennighoff, Soldaini, Groeneveld, Lo, Morrison, Min, Shi, Walsh, Tafjord, Lambert, et~al.]{muennighoff2024olmoe}
Niklas Muennighoff, Luca Soldaini, Dirk Groeneveld, Kyle Lo, Jacob Morrison, Sewon Min, Weijia Shi, Pete Walsh, Oyvind Tafjord, Nathan Lambert, et~al.
\newblock Olmoe: Open mixture-of-experts language models.
\newblock \emph{arXiv preprint arXiv:2409.02060}, 2024.

\bibitem[Nye et~al.(2021)Nye, Andreassen, Gur-Ari, Michalewski, Austin, Bieber, Dohan, Lewkowycz, Bosma, Luan, et~al.]{nye2021show}
Maxwell Nye, Anders~Johan Andreassen, Guy Gur-Ari, Henryk Michalewski, Jacob Austin, David Bieber, David Dohan, Aitor Lewkowycz, Maarten Bosma, David Luan, et~al.
\newblock Show your work: Scratchpads for intermediate computation with language models.
\newblock 2021.

\bibitem[{OpenAI}(2025)]{openai_gpt5}
{OpenAI}.
\newblock Gpt-5 [large language model].
\newblock \url{https://openai.com/index/introducing-gpt-5/}, 2025.
\newblock Released August 7, 2025.

\bibitem[OpenAI et~al.(2025)OpenAI, :, Agarwal, Ahmad, Ai, Altman, Applebaum, Arbus, Arora, Bai, Baker, Bao, Barak, Bennett, Bertao, Brett, Brevdo, Brockman, Bubeck, Chang, Chen, Chen, Cheung, Clark, Cook, Dukhan, Dvorak, Fives, Fomenko, Garipov, Georgiev, Glaese, Gogineni, Goucher, Gross, Guzman, Hallman, Hehir, Heidecke, Helyar, Hu, Huet, Huh, Jain, Johnson, Koch, Kofman, Kundel, Kwon, Kyrylov, Le, Leclerc, Lennon, Lessans, Lezcano-Casado, Li, Li, Lin, Liss, Lily, Liu, Liu, Lu, Lu, Martinovic, McCallum, McGrath, McKinney, McLaughlin, Mei, Mostovoy, Mu, Myles, Neitz, Nichol, Pachocki, Paino, Palmie, Pantuliano, Parascandolo, Park, Pathak, Paz, Peran, Pimenov, Pokrass, Proehl, Qiu, Raila, Raso, Ren, Richardson, Robinson, Rotsted, Salman, Sanjeev, Schwarzer, Sculley, Sikchi, Simon, Singhal, Song, Stuckey, Sun, Tillet, Toizer, Tsimpourlas, Vyas, Wallace, Wang, Wang, Watkins, Weil, Wendling, Whinnery, Whitney, Wong, Yang, Yang, Yasunaga, Ying, Zaremba, Zhan, Zhang, Zhang, Zhang, and
  Zhao]{openai2025gptoss120bgptoss20bmodel}
OpenAI, :, Sandhini Agarwal, Lama Ahmad, Jason Ai, Sam Altman, Andy Applebaum, Edwin Arbus, Rahul~K. Arora, Yu~Bai, Bowen Baker, Haiming Bao, Boaz Barak, Ally Bennett, Tyler Bertao, Nivedita Brett, Eugene Brevdo, Greg Brockman, Sebastien Bubeck, Che Chang, Kai Chen, Mark Chen, Enoch Cheung, Aidan Clark, Dan Cook, Marat Dukhan, Casey Dvorak, Kevin Fives, Vlad Fomenko, Timur Garipov, Kristian Georgiev, Mia Glaese, Tarun Gogineni, Adam Goucher, Lukas Gross, Katia~Gil Guzman, John Hallman, Jackie Hehir, Johannes Heidecke, Alec Helyar, Haitang Hu, Romain Huet, Jacob Huh, Saachi Jain, Zach Johnson, Chris Koch, Irina Kofman, Dominik Kundel, Jason Kwon, Volodymyr Kyrylov, Elaine~Ya Le, Guillaume Leclerc, James~Park Lennon, Scott Lessans, Mario Lezcano-Casado, Yuanzhi Li, Zhuohan Li, Ji~Lin, Jordan Liss, Lily, Liu, Jiancheng Liu, Kevin Lu, Chris Lu, Zoran Martinovic, Lindsay McCallum, Josh McGrath, Scott McKinney, Aidan McLaughlin, Song Mei, Steve Mostovoy, Tong Mu, Gideon Myles, Alexander Neitz, Alex Nichol, Jakub
  Pachocki, Alex Paino, Dana Palmie, Ashley Pantuliano, Giambattista Parascandolo, Jongsoo Park, Leher Pathak, Carolina Paz, Ludovic Peran, Dmitry Pimenov, Michelle Pokrass, Elizabeth Proehl, Huida Qiu, Gaby Raila, Filippo Raso, Hongyu Ren, Kimmy Richardson, David Robinson, Bob Rotsted, Hadi Salman, Suvansh Sanjeev, Max Schwarzer, D.~Sculley, Harshit Sikchi, Kendal Simon, Karan Singhal, Yang Song, Dane Stuckey, Zhiqing Sun, Philippe Tillet, Sam Toizer, Foivos Tsimpourlas, Nikhil Vyas, Eric Wallace, Xin Wang, Miles Wang, Olivia Watkins, Kevin Weil, Amy Wendling, Kevin Whinnery, Cedric Whitney, Hannah Wong, Lin Yang, Yu~Yang, Michihiro Yasunaga, Kristen Ying, Wojciech Zaremba, Wenting Zhan, Cyril Zhang, Brian Zhang, Eddie Zhang, and Shengjia Zhao.
\newblock gpt-oss-120b \& gpt-oss-20b model card, 2025.
\newblock URL \url{https://arxiv.org/abs/2508.10925}.

\bibitem[Patel et~al.(2021)Patel, Bhattamishra, and Goyal]{patel2021svamp}
Arkil Patel, Satwik Bhattamishra, and Navin Goyal.
\newblock Are nlp models really able to solve simple math word problems?
\newblock \emph{arXiv preprint arXiv:2103.07191}, 2021.

\bibitem[Sap et~al.(2019)Sap, Rashkin, Chen, LeBras, and Choi]{sap2019socialiqa}
Maarten Sap, Hannah Rashkin, Derek Chen, Ronan LeBras, and Yejin Choi.
\newblock Socialiqa: Commonsense reasoning about social interactions.
\newblock \emph{arXiv preprint arXiv:1904.09728}, 2019.

\bibitem[Shazeer et~al.(2017)Shazeer, Mirhoseini, Maziarz, Davis, Le, Hinton, and Dean]{shazeer2017outrageously}
Noam Shazeer, Azalia Mirhoseini, Krzysztof Maziarz, Andy Davis, Quoc Le, Geoffrey Hinton, and Jeff Dean.
\newblock Outrageously large neural networks: The sparsely-gated mixture-of-experts layer.
\newblock \emph{arXiv preprint arXiv:1701.06538}, 2017.

\bibitem[Shelmanov et~al.(2021)Shelmanov, Tsymbalov, Puzyrev, Fedyanin, Panchenko, and Panov]{shelmanov2021certain}
Artem Shelmanov, Evgenii Tsymbalov, Dmitri Puzyrev, Kirill Fedyanin, Alexander Panchenko, and Maxim Panov.
\newblock How certain is your transformer?
\newblock In \emph{Proceedings of the 16th Conference of the European Chapter of the Association for Computational Linguistics: Main Volume}, pp.\  1833--1840, 2021.

\bibitem[Taubenfeld et~al.(2025)Taubenfeld, Sheffer, Ofek, Feder, Goldstein, Gekhman, and Yona]{taubenfeld2025confidence}
Amir Taubenfeld, Tom Sheffer, Eran Ofek, Amir Feder, Ariel Goldstein, Zorik Gekhman, and Gal Yona.
\newblock Confidence improves self-consistency in llms.
\newblock \emph{arXiv preprint arXiv:2502.06233}, 2025.

\bibitem[Wang et~al.(2024)Wang, Prasad, Stengel-Eskin, and Bansal]{wang2024soft}
Han Wang, Archiki Prasad, Elias Stengel-Eskin, and Mohit Bansal.
\newblock Soft self-consistency improves language model agents.
\newblock \emph{arXiv preprint arXiv:2402.13212}, 2024.

\bibitem[Wang et~al.(2022)Wang, Wei, Schuurmans, Le, Chi, Narang, Chowdhery, and Zhou]{wang2022self}
Xuezhi Wang, Jason Wei, Dale Schuurmans, Quoc Le, Ed~Chi, Sharan Narang, Aakanksha Chowdhery, and Denny Zhou.
\newblock Self-consistency improves chain of thought reasoning in language models.
\newblock \emph{arXiv preprint arXiv:2203.11171}, 2022.

\bibitem[Watanabe(2023)]{watanabe2023tree}
Shuhei Watanabe.
\newblock Tree-structured parzen estimator: Understanding its algorithm components and their roles for better empirical performance.
\newblock \emph{arXiv preprint arXiv:2304.11127}, 2023.

\bibitem[Wei et~al.(2022)Wei, Wang, Schuurmans, Bosma, Xia, Chi, Le, Zhou, et~al.]{wei2022chain}
Jason Wei, Xuezhi Wang, Dale Schuurmans, Maarten Bosma, Fei Xia, Ed~Chi, Quoc~V Le, Denny Zhou, et~al.
\newblock Chain-of-thought prompting elicits reasoning in large language models.
\newblock \emph{Advances in neural information processing systems}, 35:\penalty0 24824--24837, 2022.

\bibitem[Yao et~al.(2023)Yao, Yu, Zhao, Shafran, Griffiths, Cao, and Narasimhan]{yao2023tree}
Shunyu Yao, Dian Yu, Jeffrey Zhao, Izhak Shafran, Tom Griffiths, Yuan Cao, and Karthik Narasimhan.
\newblock Tree of thoughts: Deliberate problem solving with large language models.
\newblock \emph{Advances in neural information processing systems}, 36:\penalty0 11809--11822, 2023.

\bibitem[Zellers et~al.(2019)Zellers, Holtzman, Bisk, Farhadi, and Choi]{zellers2019hellaswag}
Rowan Zellers, Ari Holtzman, Yonatan Bisk, Ali Farhadi, and Yejin Choi.
\newblock Hellaswag: Can a machine really finish your sentence?
\newblock \emph{arXiv preprint arXiv:1905.07830}, 2019.

\end{thebibliography}
\bibliographystyle{iclr2025_conference}

\newpage

\appendix

\section{Experiment Setup Details}

\begin{table*}[h!]
\centering
\small % Use a smaller font size to help the table fit within the column width
\caption{Experiment setup for model tuning across different task categories.
For each category, we specify the tuning configuration and model-specific hyperparameters. $N$ denotes the number of RoE samples, $T$ is the maximum temperature in tuning, and $L$ refers to the number of initial and final skipped layers. Sample denotes the number of validation set samples, Trial is the number of optimization trials. PPL denotes if perplexity was used as the optimization objective.}
\label{tab:experiment-setup}
% Reduce inter-column spacing to help the table fit. The default is 6pt.
\setlength{\tabcolsep}{4pt} 
\begin{tabular}{llc c ccc ccc ccc}
\toprule
% --- Table Header ---
% Row 1: Main column titles and model super-columns
\multirow{2}{*}{Category} & \multirow{2}{*}{Task} & \multirow{2}{*}{Sample/Trial} & \multirow{2}{*}{PPL} & \multicolumn{3}{c}{OLMoE} & \multicolumn{3}{c}{Mixtral} & \multicolumn{3}{c}{GPT-OSS} \\
\cmidrule(lr){5-7} \cmidrule(lr){8-10} \cmidrule(lr){11-13}
% Row 2: Sub-column titles for the models
& & & & $N$ & $T$ & $L$ & $N$ & $T$ & $L$ & $N$ & $T$ & $L$ \\
\midrule

% --- Math Category ---
\multirow{5}{*}{Math} & GSM8K    & \multirow{5}{*}{100/50} & \multirow{5}{*}{\checkmark} & \multirow{5}{*}{32} & \multirow{5}{*}{0.5} & \multirow{5}{*}{1} & \multirow{5}{*}{64} & \multirow{5}{*}{0.25} & \multirow{5}{*}{5} & \multirow{5}{*}{64} & \multirow{5}{*}{0.2} & \multirow{5}{*}{5} \\
& AddSub     & & & & & & & & & & & \\
& SVAMP     & & & & & & & & & & & \\
& MultiArith  & & & & & & & & & & & \\
& SingleEq    & & & & & & & & & & & \\
\midrule

% --- Commonsense Category ---
\multirow{5}{*}{Common} & SiQA        & \multirow{5}{*}{300/100} & \multirow{5}{*}{\ding{55}} & \multirow{5}{*}{32} & \multirow{5}{*}{0.5} & \multirow{5}{*}{3} & \multirow{5}{*}{64} & \multirow{5}{*}{0.3} & \multirow{5}{*}{3} & \multirow{5}{*}{64} & \multirow{5}{*}{0.2} & \multirow{5}{*}{5} \\
& OBQA        & & & & & & & & & & & \\
& Hellaswag   & & & & & & & & & & & \\
& ARC-E       & & & & & & & & & & & \\
& ARC-C       & & & & & & & & & & & \\
\midrule

% --- Code Category ---
\multirow{2}{*}{Code} & Humaneval    & \multirow{2}{*}{50/50} & \multirow{2}{*}{\ding{55}} & \multirow{2}{*}{32} & \multirow{2}{*}{0.5} & \multirow{2}{*}{1} & \multirow{2}{*}{64} & \multirow{2}{*}{0.25} & \multirow{2}{*}{5} & \multirow{2}{*}{64} & \multirow{2}{*}{0.2} & \multirow{2}{*}{5} \\
& Humaneval+) & & & & & & & & & & & \\
\bottomrule
\end{tabular}
\end{table*}

\newpage
\section{Impact of Routing Temperature}
\label{app:temp_analysis}

The routing temperature is a key hyperparameter in RoE, governing the diversity of sampled expert paths. To understand its effect, we conduct a sensitivity analysis where we apply a uniform temperature across all MoE layers and sweep its value in increments of 0.05.

\begin{figure}[h]
    \centering
    \includegraphics[]{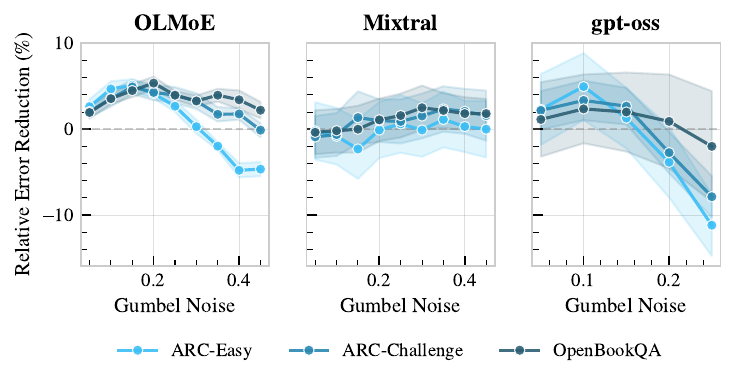}
    \caption{
        Impact of routing temperature on task performance. We apply a uniform temperature across all MoE layers and observe a concave relationship where performance peaks at a task-specific optimal value. Excessively high temperatures degrade performance by introducing noise into the expert selection process.
    }
    \label{fig:temp_sweep}
\end{figure}

As shown in Figure~\ref{fig:temp_sweep}, performance consistently follows a concave trend: accuracy improves as the temperature increases from zero, peaks at an optimal value, and then declines. This decline occurs because excessively high temperatures introduce too much noise into the routing decisions, leading to the selection of less relevant experts and degrading the final prediction quality. Crucially, we observe that the optimal temperature is task-specific, which underscores the importance of tuning this hyperparameter for each downstream application to maximize performance gains.

\newpage
\section{Tuning History and Results Heatmap}

\begin{figure}[h]
    \centering
    \includegraphics[]{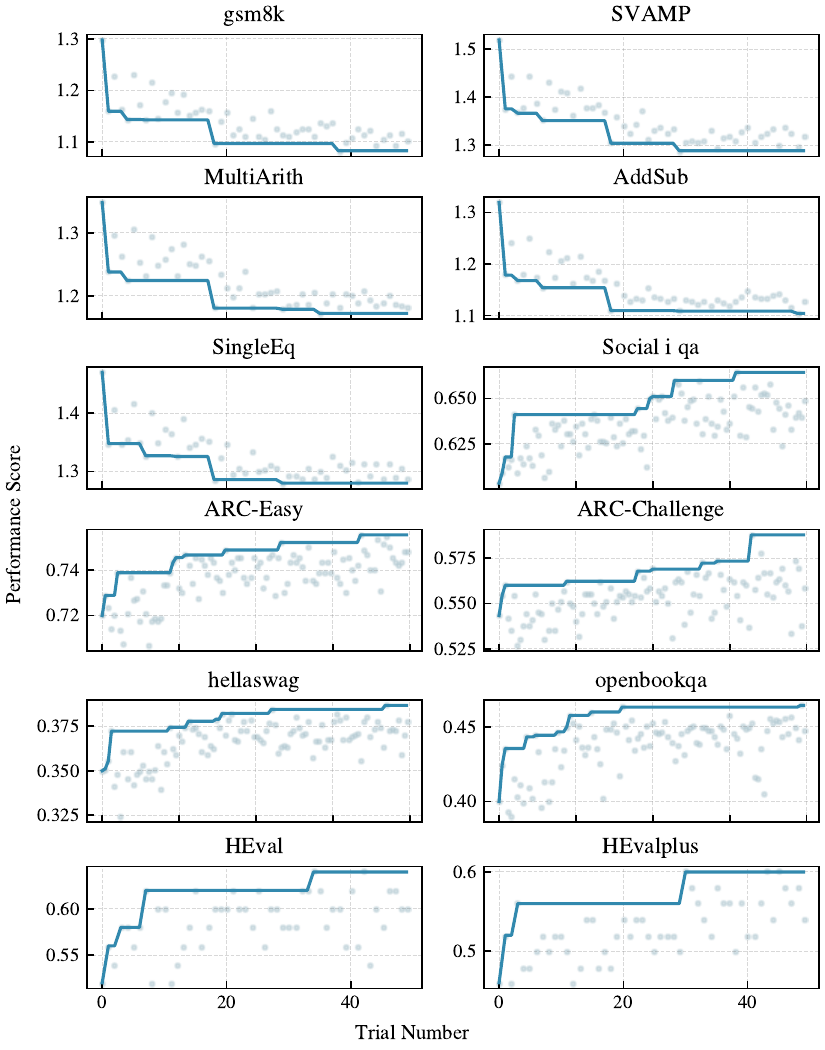}
    \caption{
        Optimization history of OLMoE on the benchmarks on the validation set. Test set evaluation results are provided in Figure~\ref{fig:main_result}
    }
    \label{fig:opt_history_olmoe}
\end{figure}

\begin{figure}[h]
    \centering
    \includegraphics[]{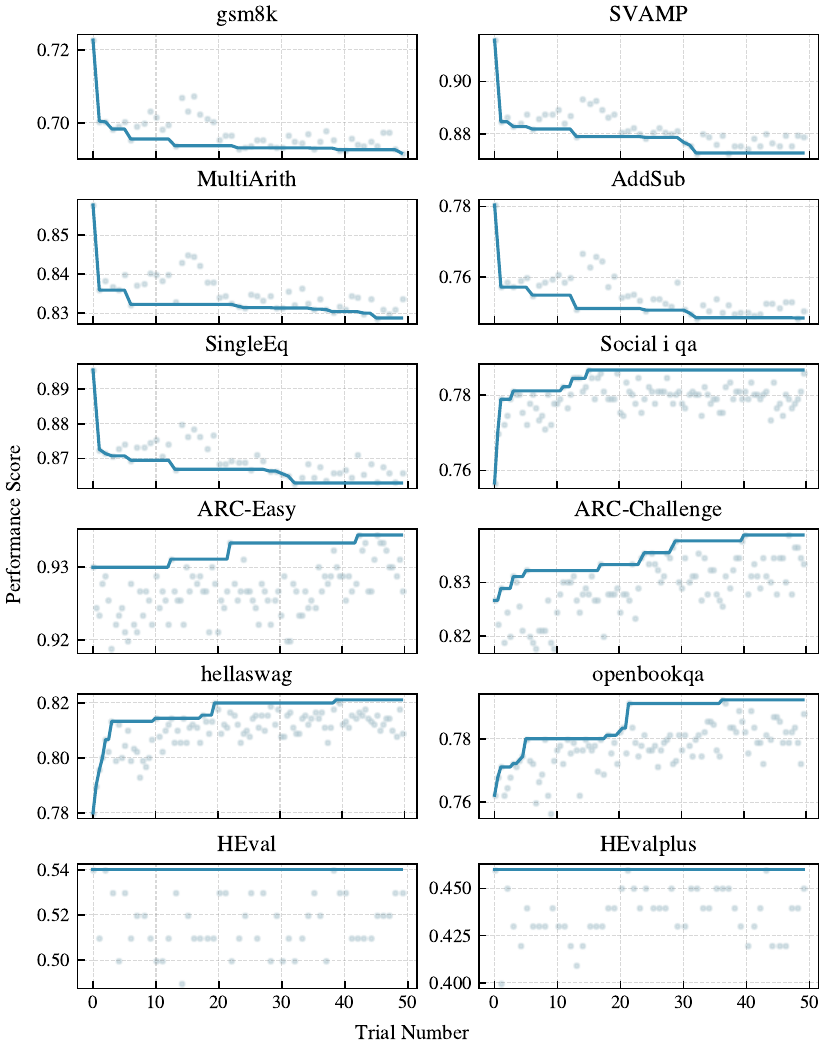}
    \caption{
        Optimization history of Mixtral on the benchmarks on the validation set. Test set evaluation results are provided in Figure~\ref{fig:main_result}
    }
    \label{fig:opt_history_mixral}
\end{figure}

\begin{figure}[h]
    \centering
    \includegraphics[]{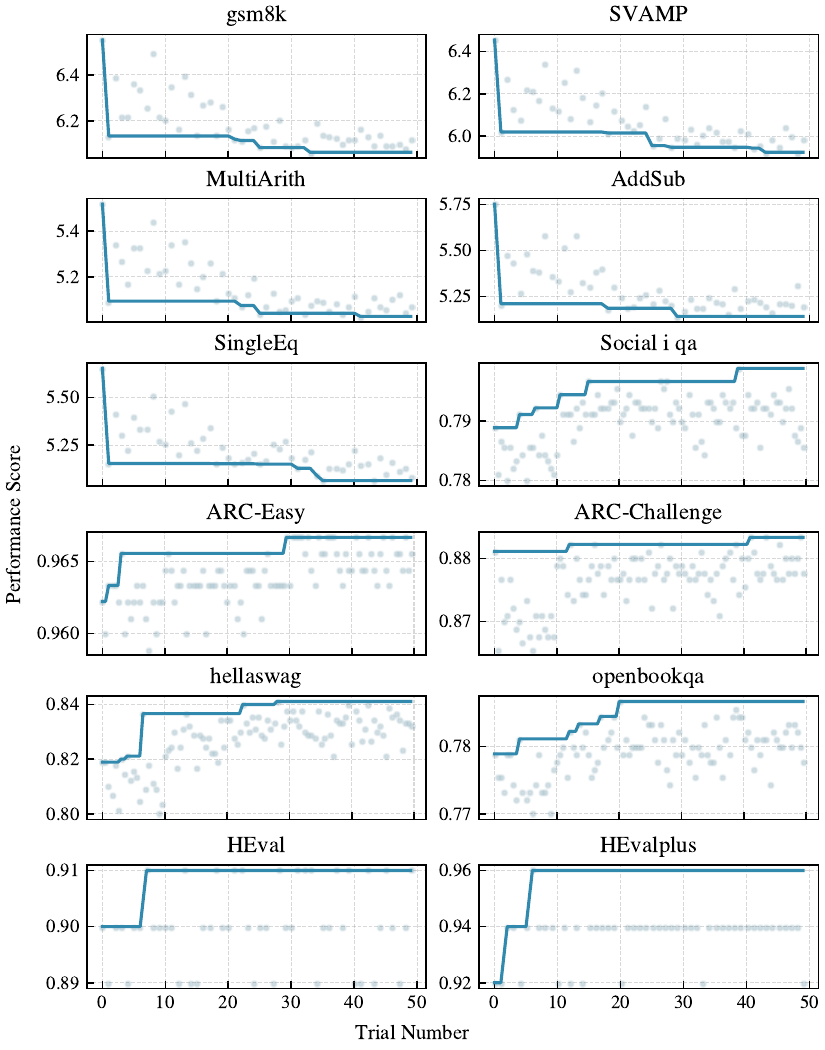}
    \caption{
        Optimization history of GPT-OSS on the benchmarks on the validation set. Test set evaluation results are provided in Figure~\ref{fig:main_result}
    }
    \label{fig:opt_history_oss}
\end{figure}

\begin{figure}[h!]
    \centering
    % --- First Subfigure ---
    \begin{subfigure}[b]{0.8\linewidth}
        \includegraphics[width=\linewidth]{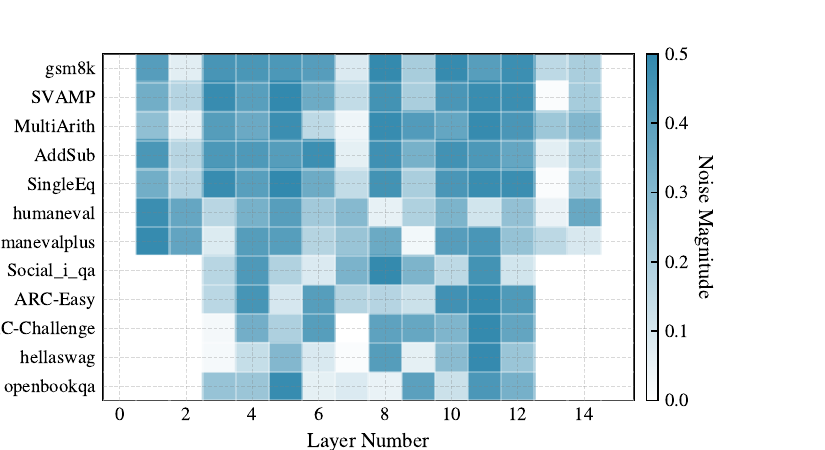}
        \caption{OLMoE-1b-7b layer temperature heatmap.}
        \label{fig:sub1}
    \end{subfigure}
    \\ % This command creates the vertical space between figures
    % --- Second Subfigure ---
    \begin{subfigure}[b]{0.8\linewidth}
        \includegraphics[width=\linewidth]{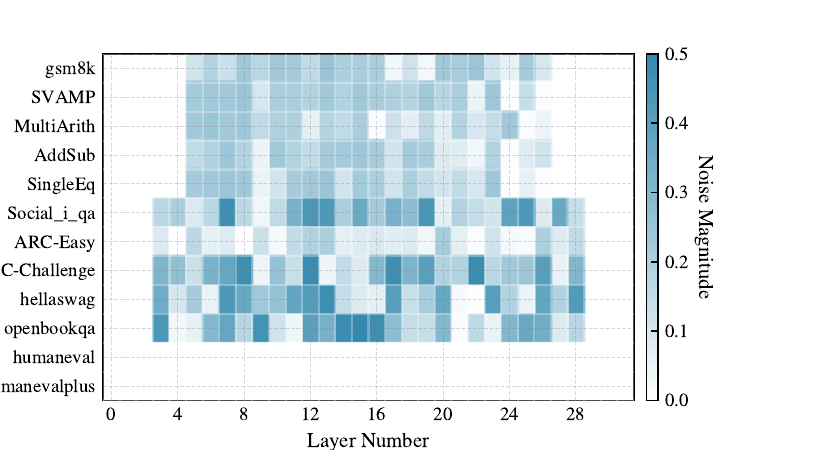}
        \caption{Mixtral-7x8 layer temperature heatmap.}
        \label{fig:sub2}
    \end{subfigure}
    \\ % This command creates the vertical space between figures
    % --- Third Subfigure ---
    \begin{subfigure}[b]{0.8\linewidth}
        \includegraphics[width=\linewidth]{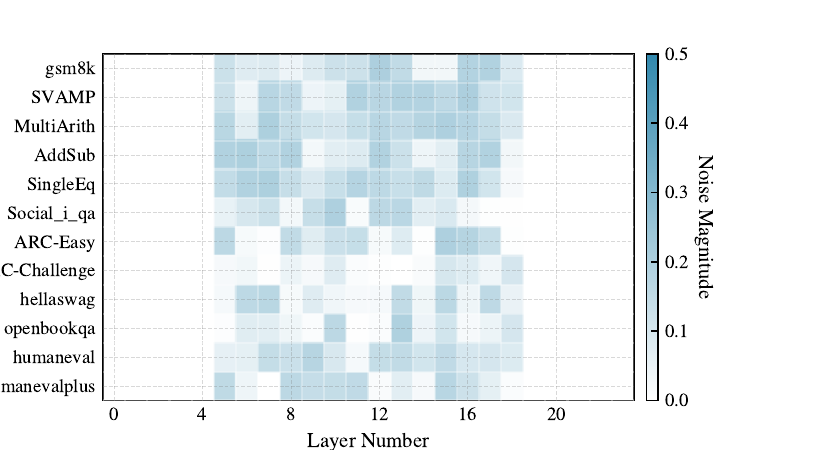}
        \caption{GPT-OSS-20B layer temperature heatmap.}
        \label{fig:sub3}
    \end{subfigure}
    
    % --- Main Caption and Label ---
    \caption{Heatmap of per layer temperature of OLMoE, Mixtral, and GPT-OSS after hyper-parameter tuning on different layers.}
    \label{fig:heatmap}
\end{figure}

% \appendix
% \section{Appendix}

\end{document}